# AgroLLM: Connecting Farmers and Agricultural Practices through Large Language Models for Enhanced Knowledge Transfer and Practical Application

Dinesh Jackson Samuel[1], Inna Skarga-Bandurova[2], David Sikolia[3], Muhammad Awais[4]

[1] Mathematics, Pittsburg State University, United States.

[2] School of Engineering, Computing and Mathematics, Oxford Brookes University, United Kingdom

[3] Kelce College of Business, Pittsburg State University, United States

[4] Department of Medical Physics, MSK Cancer Center, United States

**Abstract.**

Large Language Models (LLMs) are revolutionizing agriculture by improving knowledge sharing, precision farming, and decision-making. These models provide educational resources, interactive tutorials, and multilingual support, ensuring that farmers in diverse regions have access to critical information. The development of Agriculture LLM (AgroLLM), an innovative chatbot, has been designed to enhance education and knowledge-sharing in the agricultural sector, using AI to transform domain-specific learning. Utilizing Large Language Models (LLMs) and a Retrieval Augmented Generation (RAG) framework, AgroLLM delivers accurate, contextually relevant responses derived from a comprehensive corpus of open-source agricultural resources, including textbooks, research articles, and blogs. RAG reduces incorrect responses to agriculture related queries by retrieving relevant information from the dataset. We conducted a comparative study of three state-of-the-art models: Gemini 1.5 Flash, ChatGPT-4o Mini, and Mistral-7B-Instruct-v0.2. For the evaluation of the models, we classified the agricultural data into four key fields: Agriculture and Life Sciences, Agricultural Management, Agriculture and Forestry, and Agriculture Business. AgroLLM utilizes the Facebook AI Similarity Search (FAISS) vector database for efficient similarity searches and retrieval, ensuring rapid and precise access to agricultural knowledge. Each model was assessed based on several key parameters: the quality of the embeddings they produced, the efficiency of their similarity searches, and the coherence and relevance of the responses generated by the LLM. Finally, a comparative study of three state-of-the-art models embedded in the AgroLLM framework like Gemini 1.5 Flash, ChatGPT-4o-mini, and Mistral-7B-Instruct-v0.2 was conducted to evaluate performance, response time, and contextual relevance for complex agricultural queries. The results showed that ChatGPT-4o mini with RAG outperforms the alternatives, with an accuracy of 93%. Continuous feedback mechanisms are incorporated to refine response quality, adapt to evolving user needs, and maintain high accuracy over time. This approach not only supports effective education and research in agriculture but also establishes a new benchmark for AI-driven educational tools, empowering learners and professionals in the agricultural domain

## 1. Introduction

In recent years, Large Language Models (LLMs) have made significant strides, becoming a driver for many innovative solutions across different areas. These models are designed to learn and generate human-like responses based on the context and data they are trained on, utilizing reinforcement learning from human feedback (Kasneci et al., 2023; Zhao et al., 2023). By identifying patterns and semantic relationships in text, LLMs can generate accurate and coherent responses akin to human communication. Their capabilities now extend beyond text generation to creating image captions, writing code snippets, assisting users virtually, performing data analysis, and more. With access to diverse datasets, LLMs have become transformative tools in sectors such as healthcare (Cascella et al., 2023), robotics (Wang et al., 2024), agriculture (Marinoudi et al., 2024), hydrology (Samuel et al., 2024a, Samuel et al., 2024b), and road and traffic safety (Zarza et al., 2023). The evolution of LLMs began with Recurrent Neural Networks (RNNs) for processing sequential text and progressed to more advanced architectures like Long Short-Term Memory (LSTM) networks and Transformer models. OpenAI's breakthrough Generative Pre-training Transformer (GPT) architecture significantly improved the handling of long-term dependencies in text (OpenAI, 2024), establishing a robust foundation for LLMs. Following GPT's success, notable advancements include DeepMind's dialogue-focused LaMDA (Thoppilan et al., 2022) and Google's Gemini (Gemini, 2025).

The ability of LLMs to understand queries and generate human-like responses, providing personalized recommendations make them key players in the development of more intuitive, conversational interfaces. AI-powered chatbots have facilitated human-like interactions in a more conversational manner (Lin et al., 2023, Khosravi et al., 2024, Alazzam et al., 2023). This human-like intelligence is achieved through advancements in high-computation systems, vast amounts of data, and improved performance of LLMs (Zhao et al., 2023, Koubaa et al., 2023). Consequently, chatbots have become increasingly popular and are being developed for various fields. These chatbots can understand human language and respond with accurate information in a way that is easily understandable, making LLMs essential tools in healthcare (Sallam et al., 2023, Ayanouz et al., 2020, Athota et al., 2020), research (Aydın 2023, Macdonald et al., 2023, Girotra 2023), and education (Sajja et al., 2023). Before the advent of LLMs, early chatbots were limited in contextual understanding and domain specificity. They also faced scalability issues when deployed across various platforms. However, the release of ChatGPT 3.5 by OpenAI in November 2022 marked a significant advancement in chatbot technology (Bahrini et al., 2023, Zhang et al., 2023). This was followed by the introduction of GPT-4 (also known as ChatGPT Plus (Achiam 2023), which further demonstrated the capabilities of human-like conversational chatbots. Following OpenAI's innovations, Google released BARD, the first LLM-based chatbot (Ortiz 2023). In February 2024, Google DeepMind released Gemini AI, which became the successor to LaMDA and PaLM 2 (Saeidnia, 2023, Yllemo, 2023). The rapid evolution of AI-based conversational chatbots has made them an excellent addition to knowledge-based systems, expanding their applications across various sectors.

Agriculture is one of the most crucial and, at the same time, promising sectors where AI technologies are making a significant impact. Agriculturists ensure proper resource utilization while boosting crop production and awareness regarding environmental-friendly practices. They fight against climate change issues, create protective boundaries for ecosystems, and help enhance agriculture progress to meet the needs of the future. In this aspect, AI-driving agriculture is rapidly gaining momentum as a vital tool for improving farming practices and promoting sustainability. A wide variety of AI-based approaches have been implemented for plant disease diagnosis (Zhang, 2023), pest control management (Anticimex, 2021, Kariyanna and Sowjanya, 2024), crop monitoring (Kaya, 2025), boosting crop yields (Kamilaris et al., 2018; Tzachor et al., 2023) and many others, fostering education and decision-making processes.

To utilize the full potential of AI, farmers should be equipped with a mixed skill set, including technical know-how, analytical thinking, and adaptability to new technologies. For this reason, there is a high demand for enhanced agricultural education, and LLMs become the emerging tools in this educational shift, bridging cutting-edge agricultural science with day-to-day farming practices (Peng et al., 2023; Zhu et al., 2024). Another important aspect of using LLMs in agriculture is their potential to be used with IoT systems and their ability to consolidate, structure and simplify the navigation of agricultural knowledge, assisting farmers with their everyday tasks. Autonomous chatbots provide 24/7 assistance for multi-featured problems and queries related to planting schedules, farm management and resource allocation, offering solutions tailored to the specific conditions of each farm.

In spite of big promises, utilizing LLMs for agriculture faces several significant challenges related to the domain and knowledge specifics. Thus, for example, agricultural science utilizes specialized vocabulary and concepts (e.g., soil nutrient cycles, integrated pest management) that generic LLMs may not fully capture unless they are fine-tuned on agriculture-specific datasets. Accurate understanding of context-specific information (like regional farming practices or environmental factors) requires domain adaptation, which may be challenging given the high diversity within agricultural practices around the world. Data quality and availability raise another challenge because agricultural data come from different sources (textbooks, research articles, field reports, sensor data) and vary in quality, formats, and granularity. There is often a scarcity of large, high-quality, and annotated datasets that cover niche agricultural topics compared to more generalized datasets in other domains (e.g., general web data or news). Finally, when an LLM is applied to specialized domains like agriculture, errors or misinterpretations in the generated content may lead to practical problems, such as incorrect advice on crop management or resource allocation. As a result, farmers and agricultural experts may be skeptical of automated systems, especially if the outputs are overly generic.

This project aims to develop a comprehensive agriculture-specific LLM (AgroLLM) as a centralized knowledge base LLM for agriculture. As the first step to achieving this goal, this paper provides a structured approach to integrating LLMs into agricultural applications, demonstrating the entire pipeline, from data collection and preprocessing to embedding generation, retrieval using vector databases, and retrieval-augmented generation. It shows how existing technologies can be adapted to a specialized domain, serving as a proof of concept that can spur further research and development. Our contribution to this work includes the following:

1. The AgroLLM bridges the knowledge gap by connecting theoretical knowledge with practical applicability for farmers.

2. We used an agriculture-specific database using an extensive corpus of university-referenced textbooks, open source web contents and research articles to facilitate efficient information retrieval in the agricultural domain.
3. We designed the AgroLLM framework for agricultural knowledge generation, which uses a combination of cutting-edge AI technologies to deliver contextually relevant and accurate agricultural information. The text extraction is performed using generative AI, and the resulting embeddings are stored in Facebook AI Similarity Search (FAISS) format. This ensures efficient similarity searches and retrieval along with precise access to agricultural knowledge.
4. To select benchmark generative AI model for the AgroLLM, we conducted a comparative analysis of three LLM models: Gemini 1.5 Flash, ChatGPT-4o-mini, and Mistral-7B-Instruct-v0.2, and assessed their effectiveness in generating responses related to farming and agriculture. The evaluation parameters include accuracy, relevance, response time, and ability of models to respond complex agricultural queries.

## 2. Methodology

Overall methodology described in this section describes three main components (1) data collection and processing, (2) AgroLLM framework, and (3) LLM evaluation and selection.

### 2.1 Structured database for AgroLLM

The data for this project comprises a collection of the agricultural textbooks commonly referenced in universities and research articles which is divided into four topic-related groups, including:

- Agriculture and life sciences (agricultural policy studies, agricultural law, communication, and emerging research in life sciences)
- Agricultural management (agribusiness finance, product development, production management, land and resource allocation, and leadership in agricultural enterprises)
- Agriculture and forestry (crop and soil science, horticulture, and rangeland management)
- Agriculture business (ecological restoration, dairy science, organic farming, animal science, and insights into Native American agricultural practices)

To transform raw textual data into a structured, searchable database a systematic approach was used and included text extraction, annotation, categorization and statistical analysis. Annotation was performed using both manual and semi-automated procedures. The process included annotation of key concepts, methodologies, case studies, and terminologies across all defined agricultural sub-domains and linking topics to corresponding chapters where critical information was discussed. Then, based on the annotated content, the data were divided into categories and sub-categories. For instance, topics like "Crop and soil science" and "Agricultural communication" were segregated to enable focused text generation and inference. In the final step, the texts were analysed by extracting statistical metrics such as page count, word count, and frequency of key terms. The process of transforming agricultural data into actionable insights is shown in Fig. 1.

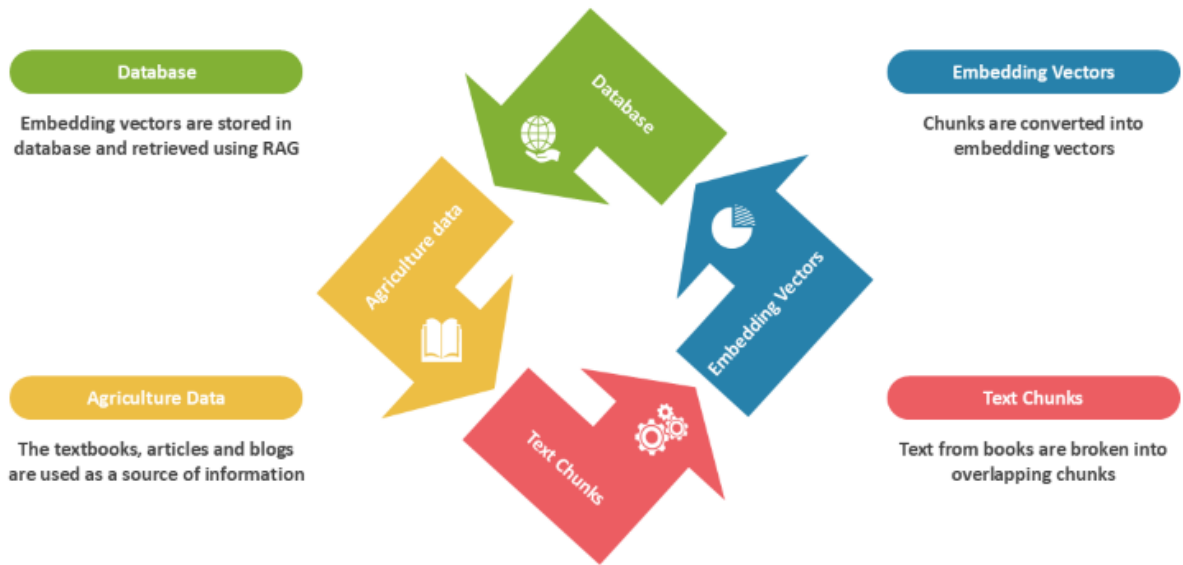

Figure 1: The process of transforming agricultural data into embedding vectors for AgroLLM.

The process begins with collecting agricultural data, which is the primary source of information. In addition to the texts extracted from the relevant literature, the dataset included test questions from textbooks, paraphrased questions, user queries, noisy text, and external context chunks to test semantic understanding and the system's ability to retrieve and generate responses to complex requests. The text data were split into overlapping text chunks, ensuring that meaningful segments of text were available for processing. Further, these book text chunks are converted into embedding vectors for efficient search and retrieval. The text chunks are subsequently converted into embedding vectors for efficient search and retrieval. The embedding vectors are then stored in a database, where they can be retrieved using Retrieval-Augmented Generation (RAG) and provide relevant information to LLM for generating response for user queries. This cycle supports the seamless transition from raw data to structured, retrievable knowledge described below.

## 2.2 AgroLLM framework

In the AgroLLM framework, we used the idea of a combination of generative AI embedding generation with a vector database for efficient search (Monir et al., 2024). The strength of this approach lies in connecting the contextual understanding from generative AI and the robust search capabilities from libraries specifically built for vector indexing.

General workflow for information retrieval and generation encompasses query processing, embedding generation, database management, and retrieval-augmented response generation, as illustrated in Figure 2.

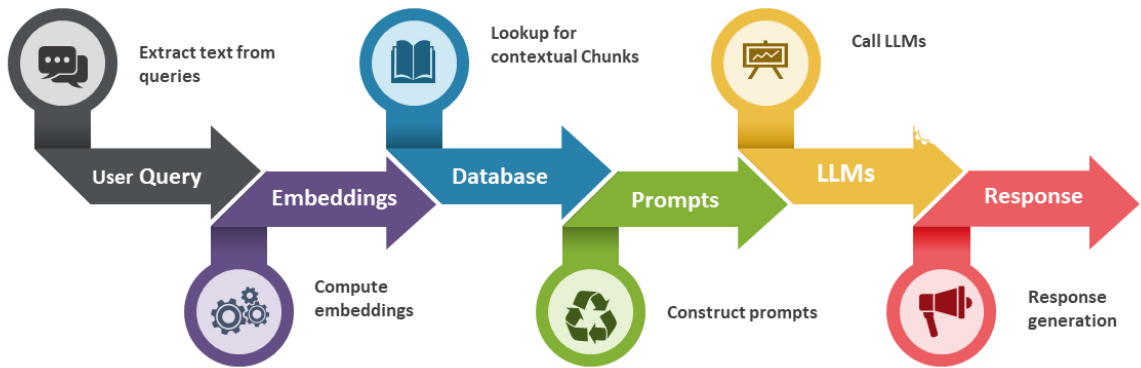

Figure 2: The workflow of AgroLLM for information generation

The process begins when a user submits a query, which then is preprocessed to extract text for embedding generation.

### 2.2.1 Embedding models: Conversion of chunks into embeddings

To manage the extensive information contained in these agriculture books, we divided the documents into smaller, manageable chunks. This chunking process is essential for creating meaningful embeddings and ensuring efficient storage and retrieval. Each chunk was carefully crafted to encapsulate a specific topic or sub-topic, ensuring that the granularity was appropriate for educational use. The chunk sizes varied depending on the complexity and cohesiveness of the content, typically ranging from a few sentences to a couple of paragraphs. We also maintained metadata for each chunk, such as chapter and section headings, to preserve context during retrieval. This metadata helps in accurately situating each chunk within the broader structure of the book, making it easier to provide relevant and contextually accurate responses when users query the system. By ensuring that each chunk is appropriately sized and contextually rich, we aim to facilitate a more effective and engaging learning experience for users.

Embeddings are dense vector representations of text that capture the semantic meaning of the chunks. We utilized embedding models from Gemini, OpenAI, and Mistral to convert our text chunks into embeddings. These models, trained on vast corpora of text, provide robust representations that facilitate effective similarity searches. Notably, Gemini and Hugging Face models use an embedding dimension of 1024, while OpenAI's model uses a dimension of 768. The higher dimension in Gemini and Hugging Face embeddings allows for a richer representation of the text but also requires more computational resources. By processing the chunks with these models, we generated embedding vectors that encapsulate the semantic context of the text. We stored the embedding vectors along with their corresponding metadata, which includes information such as chapter and section headings. This metadata is crucial for preserving context and ensuring that the retrieved information is relevant and accurately situated within the broader structure of the document. By maintaining a detailed metadata record, we enhance the capability of the chatbot to provide precise and contextually appropriate responses. This process of converting text chunks into embeddings is the crucial step in the project. It enables efficient storage and retrieval of information, allowing users to quickly access relevant and detailed agricultural knowledge. By using advanced AI models, we ensure that the embeddings are both accurate and rich in semantic content, supporting a more effective and engaging learning experience for users.

**2.2.2 Database lookup using FAISS**

The generated embeddings are stored in a vector database using the FAISS (Facebook AI Similarity Search) format. FAISS is a library designed for efficient similarity search and clustering of dense vectors, making it ideal for our extensive dataset. This step ensures that our embeddings are organized to allow rapid retrieval based on similarity measures. The choice of FAISS is driven by its scalability and performance in handling large volumes of vectors. This is crucial in the project, as it involves managing a significant amount of data derived from comprehensive agricultural texts. By implementing an indexing strategy within FAISS, we optimized the retrieval speed and accuracy. This strategy balances between exact search and approximate nearest neighbor (ANN) search, ensuring that users can quickly and accurately retrieve relevant information. Further storing embeddings in FAISS allows us to efficiently manage and query the data. When a user makes a query, the system can rapidly search through the embeddings to find the most similar chunks, providing accurate and contextually relevant responses. This setup enhances the functionality and responsiveness of our Agricultural-specific chatbot, offering improved user experience. Overall, using FAISS to store our embeddings ensures that we can handle the large-scale data efficiently while maintaining high performance in similarity searches.

**2.2.3 Prompt construction and sending relevant chunks to the LLM**

The retrieved text chunks are combined into a structured prompt, which is then sent to an LLM using the Retrieval-Augmented Generation (RAG) technique (Gao et al., 2023). The RAG approach combines text generation with searching and selecting the most relevant data from a pre-indexed vector database (Lin, 2024; Ndimbo, 2025). Figure 3 illustrates how LLM processes these chunks to generate coherent and contextually relevant responses.

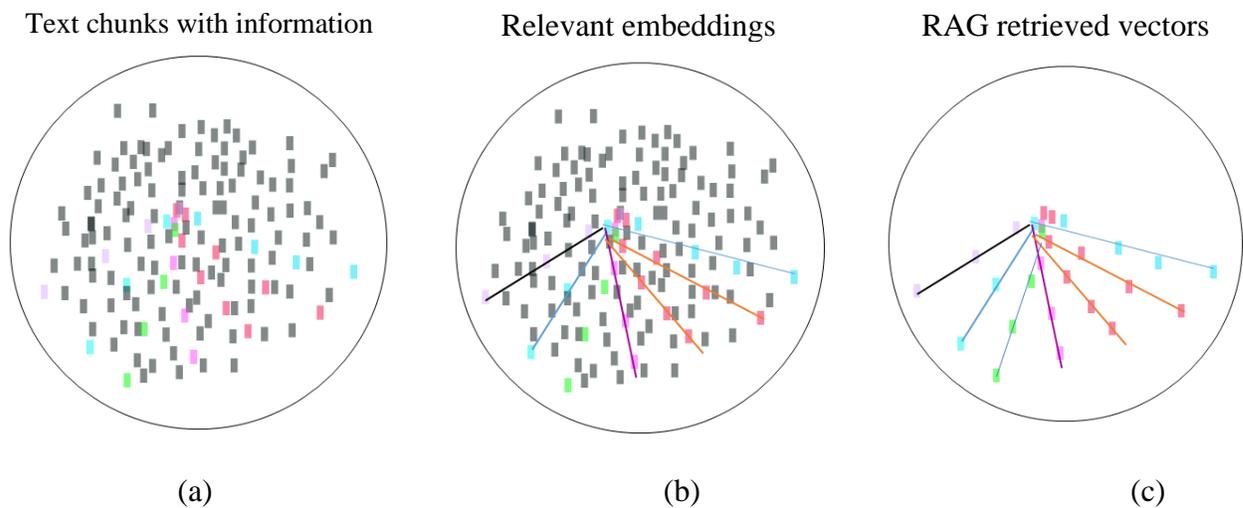

Figure 3: Illustration of the retrieval of embedding using RAG for LLMs: (a) the entire set of available data chunks retrieved from agriculture books, (b) data chunks relevant to the user query and (c) the result of the generation of relevant embeddings using RAG.

The user query and the relevant information are both given to the LLM. The LLM uses the retrieved information to create better responses. RAG allows the model to create responses based on the context obtained from the database. This reduces the risk of unpredictable or incorrect responses typical of generative models.

### 2.2.4 Response generation and knowledge sharing through AgroLLM

The AgroLLM is designed as an educational and knowledge-sharing tool tailored for agriculture-related topics and implemented as a chatbot. This bridges the knowledge gap, connecting farmers with effective agricultural practices. As a virtual assistant, AgroLLM uses the relevant text chunks to provide explanations, answer questions, and offer insights into various agricultural and farming concepts. The response generation workflow starts with a user query, which they send through the user interface, as depicted in Figure 4.

The user interface sends the query to the RAG system, which acts as the orchestrator, managing how the query is processed. The RAG converts the query into an embedding and submits it to the FAISS vector database containing pre-computed embeddings of text chunks from curated agricultural texts. The vector database performs a similarity search to identify text chunks that are semantically closest to the user query. To ensure that user receives response even when relevant documents are unavailable, the system equipped with a fallback mechanism. The RAG system evaluates whether relevant documents were retrieved. If yes, the retrieved chunks are combined with the original query to create a context-rich prompt for the LLM. Otherwise, the system falls back to the LLM's internal knowledge to generate a response. Response generation is performed through the LLM. The LLM processes the constructed prompt (including retrieved chunks and the user query) and synthesizes a detailed, coherent response. In case if no relevant documents were retrieved, the LLM uses its pre-trained internal knowledge to address the query. The generated response is sent back to the RAG system and then delivered to the user through the user interface.

The quality of the final output heavily depends on the LLM's ability to effectively integrate and articulate the information from the retrieved chunks. To select the best model for AgroLLM chatbot, we evaluated three LLMs in terms of their ability to understand and integrate information from multiple chunks.

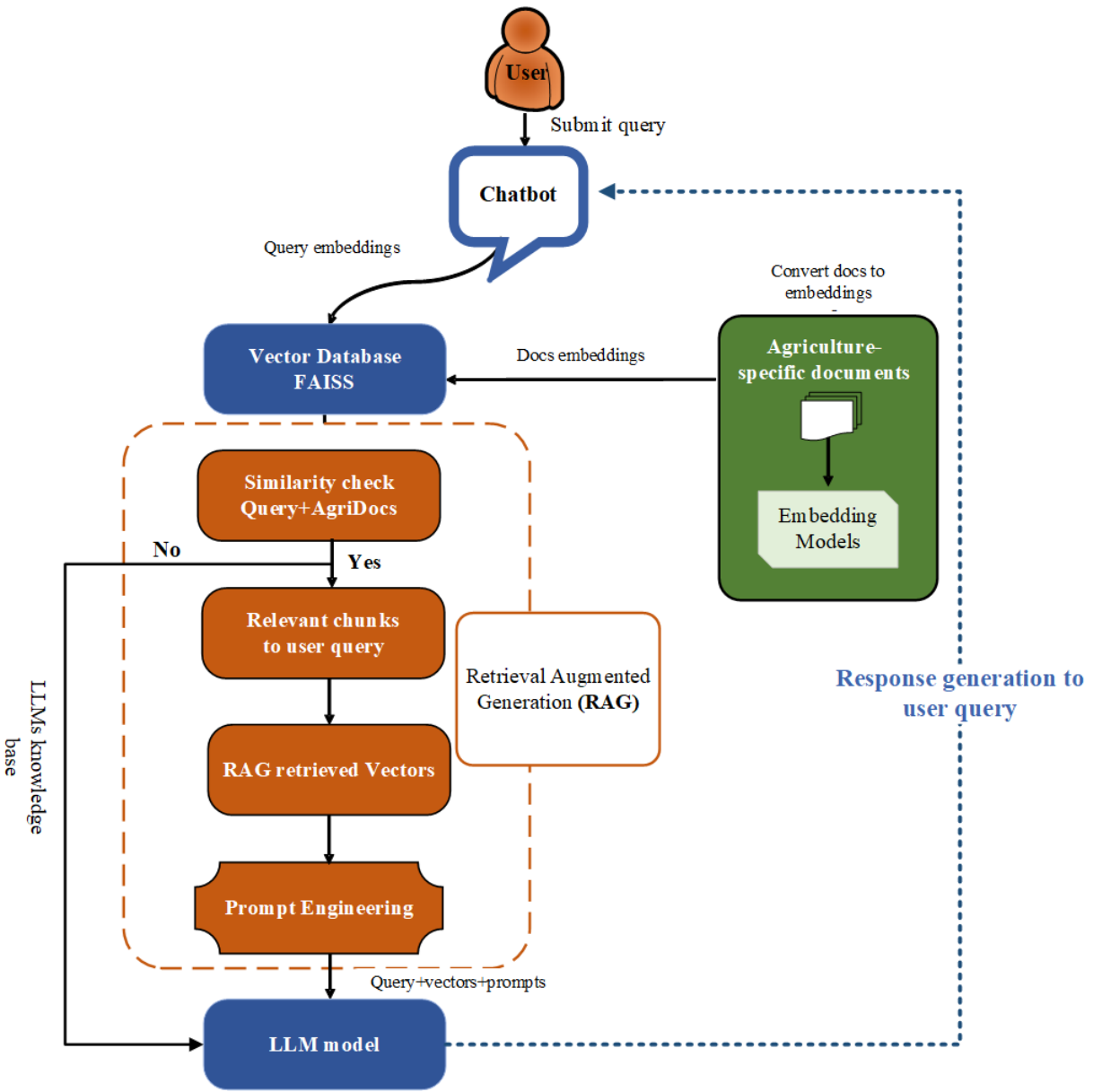

Figure 4: Workflow of AgroLLM, integrating the RAG system with a vector database and LLM

## 3. Results and Discussion

### 3.1 Evaluation parameters

For the evaluation of the models, we used the same database of agricultural textbooks described above with data ordered under four key topics: Agriculture and life sciences, Agricultural management, Agriculture and forestry, and Agriculture business, as shown in Figure 5.

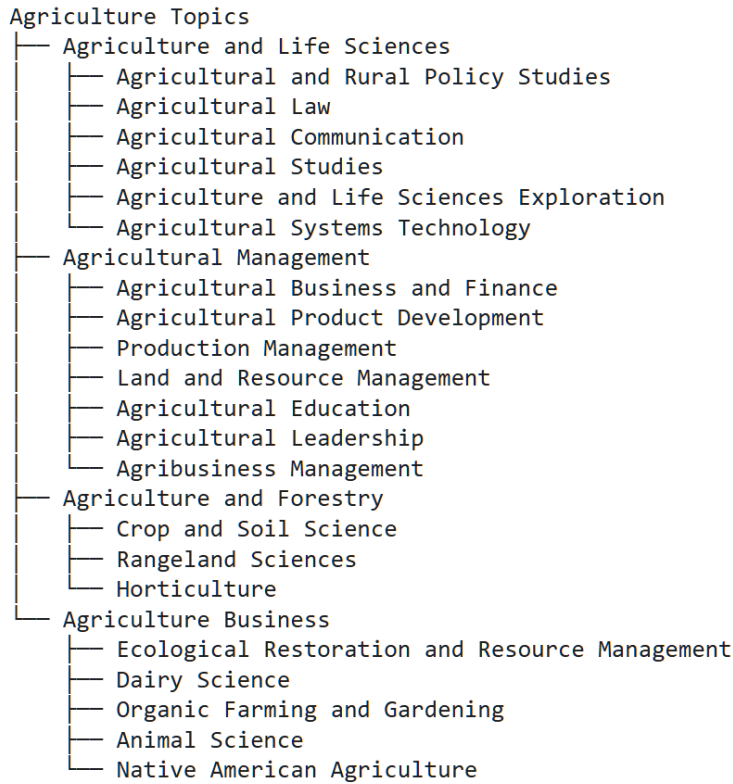

```
Agriculture Topics
├── Agriculture and Life Sciences
│   ├── Agricultural and Rural Policy Studies
│   ├── Agricultural Law
│   ├── Agricultural Communication
│   ├── Agricultural Studies
│   ├── Agriculture and Life Sciences Exploration
│   └── Agricultural Systems Technology
├── Agricultural Management
│   ├── Agricultural Business and Finance
│   ├── Agricultural Product Development
│   ├── Production Management
│   ├── Land and Resource Management
│   ├── Agricultural Education
│   ├── Agricultural Leadership
│   └── Agribusiness Management
├── Agriculture and Forestry
│   ├── Crop and Soil Science
│   ├── Rangeland Sciences
│   └── Horticulture
└── Agriculture Business
    ├── Ecological Restoration and Resource Management
    ├── Dairy Science
    ├── Organic Farming and Gardening
    ├── Animal Science
    └── Native American Agriculture
```

Figure 5: The structure of the dataset repository used as data resources for text generation.

For each topic, we selected 20 relevant questions to cover main concepts within each area. Additionally, we included questions from the book "Precision Agriculture" to challenge the models with specialized and advanced topics and estimate the relevance and accuracy of the responses generated by the models. Through this evaluation, we aimed to select the model that not only performed well across all four topics but also demonstrated best capabilities in handling specialized agriculture content. In total, 108 domain-related questions across diverse agriculture topics were generated for evaluating each model.

The relevant chunks from the agriculture data extracted by RAG were used as benchmark answers to compare the relevancy of the generated responses. In addition, a manual evaluation is performed, cross-referencing the answers with those provided in the book. This allowed us systematically assess the performance of each model, focusing on the accuracy, coherence, and relevance of their responses in the context of agricultural education.

### 3.2 The LLM comparison

We conducted a comparative study of the Mistral-7B-Instruct-v0.2 (Mistral AI), ChatGPT-4o mini (OpenAI), and Gemini 1.5 Flash (Google) models to evaluate their performance in the context of agricultural practices and education. Each model was assessed based on several key parameters: the performance, the quality of the embeddings they produced, the efficiency of their similarity searches, and the coherence and relevance of the responses generated by the LLM. From the case study, all the models were evaluated manually for their response generation capabilities. Retriever mechanism employed was FAISS with cosine similarity as similarity metric for measuring semantic relevance.

Embedding quality was evaluated using three additional metrics, (1) Mean Reciprocal Rank (MRR), which measures how well the system ranks relevant results for a query, (2) Recall@k to evaluate the proportion of relevant results retrieved within the top-k results for a set of queries, and (3) Bilingual Evaluation Understudy (BLEU) to measure how closely the AI-generated response matches a reference answers from the textbooks, capturing both coherence and relevance. For BLEU, we ccompared system-generated responses to expert-crafted and pre-annotated reference responses for agricultural questions. The results of comparison are shown in Table 1.

| Model | MRR (0-1) | Recall@10 (%) | BLEU (%) |
|---|---|---|---|
| Mistral-7B-Instruct-v0.2 (FAISS) | 0.68 | 70 | 72.12 |
| Gemini 1.5 Flash (FAISS) | 0.82 | 83 | 84.21 |
| ChatGPT-4o mini (FAISS) | 0.89 | 87 | 89.32 |
| Mistral-7B-Instruct-v0.2 (RAG) | 0.76 | 80 | 80.26 |
| Gemini 1.5 Flash (RAG) | 0.86 | 86 | 85.75 |
| ChatGPT-4o mini (RAG) | **0.93** | **92** | **91.5** |

Table 1: Embedding quality comparison in Agro-specific LLMs

The analysis of embedding quality metrics reveals that ChatGPT-4o mini with RAG consistently outperforms the other models across MRR, Recall@10, and BLEU scores, particularly when enhanced with RAG. Gemini 1.5 Flash also shows strong performance improvements with RAG, while Mistral-7B-Instruct-v0.2 exhibits the least performance, even after RAG integration.

Two key parameters, response accuracy and average response time (s) was evaluated across all topics for LLMs with FAISS and RAG-enhanced models. The results presented in Table 2 illustrate the performance of the LLMs evaluated for question answering.

| Topics | Mistral-7B-Instruct-v0.2 | | | | Gemini 1.5 Flash | | | | ChatGPT-4o mini | | | |
|---|---|---|---|---|---|---|---|---|---|---|---|---|
| | Acc. (%) | Avg. Time (s) | Acc. (%) | Avg. Time (s) | Acc. (%) | Avg. Time (s) | Acc. (%) | Avg. Time (s) | Acc. (%) | Avg. Time (s) | Acc. (%) | Avg. Time (s) |
| | FAISS | | RAG | | FAISS | | RAG | | FAISS | | RAG | |
| 1. Agriculture and life sciences | 63 | 0.5 | 67 | 29.5 | 75 | 0.3 | 80 | 4.4 | 88 | 0.2 | 92 | 11.3 |
| 2. Agricultural management | 71 | 0.5 | 74 | 27.3 | 81 | 0.4 | 85 | 4.8 | 89 | 0.4 | 93 | 9.8 |
| 3. Agriculture forestry | 72 | 0.5 | 73 | 29.2 | 83 | 0.3 | 85 | 4.8 | 91 | 0.2 | 95 | 10.7 |
| 4. Agriculture business | 69 | 0.5 | 74 | 27.5 | 82 | 0.5 | 86 | 5.2 | 92 | 0.3 | 94 | 10.6 |
| 5. Precision agriculture | 71 | 0.5 | 75 | 24.7 | 84 | 0.3 | 87 | 4.3 | 84 | 0.2 | 91 | 11.2 |
| *Average* | 69.2 | 0.5 | 72.6 | 27.64 | 81 | 0.36 | 84.6 | 4.7 | 88.8 | **0.26** | 93 | 10.72 |

Table 2: Performance comparison in Agro-specific LLMs

The results show that the RAG significantly enhances all models, but the degree of improvement varies. Table 2 compares the accuracy (%) and average response time (seconds) of different LLMs Mistral-7B-Instruct-v0.2, Gemini 1.5 Flash, and ChatGPT-4o mini across various agricultural topics using FAISS and RAG retrieval techniques. ChatGPT-4o mini consistently achieves the highest accuracy, averaging 93% with RAG and 88.8% with FAISS, while Mistral-7B has the lowest at 69.2% (FAISS) and 72.6% (RAG). Response times vary significantly, with FAISS being much faster (as low as 0.26s for ChatGPT-4o mini) compared to RAG, which takes considerably longer (10.72s for ChatGPT-4o mini). This shows that ChatGPT-4o Mini with RAG provides more accurate responses, with a trade-off in processing time.

## Conclusion

In this study, we explored the application of advanced LLMs to develop the AgroLLM chatbot, an educational tool designed to enhance practical learning and knowledge sharing in the field of Agriculture. This study presented AgroLLM, an agriculture-specific framework designed to enhance agricultural practical applicability and decision support through the integration of LLMs. By using a robust pipeline that includes data collection, semantic embedding generation, and RAG, AgroLLM serves as a centralized knowledge base tailored to the unique demands of the agricultural sector. Three language models—Mistral-7B-Instruct-v0.2, Gemini 1.5 Flash and ChatGPT-4o Mini were used and tested to see how well they work in agricultural settings. ChatGPT-4o mini was the top performer, achieving the highest accuracy (93 %) and the response time (10.72 seconds) when using RAG. ChatGPT-4o mini without RAG was very close, with an accuracy of 88.8% and a better response time. Both models showed a good balance between being accurate and fast. The findings from this study highlight the significant potential of specialized LLMs, particularly when enhanced with RAG, to serve as powerful tools in the agricultural sector. ChatGPT 4o-mini stood out as the best choice for agricultural applications. High accuracy and low response time of this model promises that AgroLLMs can be successfully integrated with IoT systems to enhance agriculture and real-time decision-making. While AgroLLM demonstrates promising capabilities, certain limitations warrant attention. First, the current framework relies on a specific set of university-referenced agricultural textbooks, research articles and web contents. Expanding the dataset to include a broader range of sources, such as field reports and real-time sensor data, could enhance the model's comprehensiveness and applicability. Finally, although ChatGPT4o-mini RAG shows superior performance, assessing its scalability across larger and more diverse datasets remains an area for future exploration. Our future research will be focused on addressing these limitations by incorporating more diverse and high-quality agricultural data, refining domain-specific training methodologies, and exploring hybrid model approaches that combine the strengths of different LLMs to further enhance performance and applicability.

## Declaration of generative AI in the writing process

During the preparation of this work, the authors used AI tools to improve the flow of the text, correct any potential grammatical errors, and improve the writing. After using the tool, authors reviewed and edited the content as needed and took full responsibility for the content of the publication.

## Data availability statement

The data that support the findings of this study are available from the corresponding author upon reasonable request.

## Conflict of interest

The authors declare that they have no conflicts of interest related to this work.